\documentclass[lettersize,journal]{IEEEtran}
\usepackage{array}
\usepackage[caption=false,font=normalsize,labelfont=sf,textfont=sf]{subfig}
\usepackage{textcomp}
\usepackage{stfloats}
\usepackage{url}
\usepackage{verbatim}
\usepackage{graphicx}
\usepackage{cite}

\usepackage{amsfonts,amsmath}
\usepackage{xparse}[2019/10/11]
\usepackage{xcolor,cleveref}

\usepackage{algorithm,algorithmicx,algpseudocode}
\usepackage{nicefrac,multirow}
\algrenewcommand\algorithmicrequire{\textbf{Input:}}
\algrenewcommand\algorithmicensure{\textbf{Output:}}

\Crefname{ALC@unique}{Line}{Lines}

\definecolor{darkgreen}{RGB}{0,128,0}
\definecolor{darkblue}{RGB}{0,0,128}
\definecolor{darkred}{RGB}{128,0,0}

\makeatletter
\newcommand{\thickhline}{%
    \noalign {\ifnum 0=`}\fi \hrule height 1pt
    \futurelet \reserved@a \@xhline
}
\makeatother

\DeclareDocumentCommand \prArg{mm}
{(
\IfNoValueTF{#2}{#1}{#1 \mid #2}
)}

\DeclareDocumentCommand \newProbabilityFormat{r<>m}
{
	\DeclareDocumentCommand #1 {e{_}e{^}>{\SplitArgument{1}{|}}r()}
	{
		\IfNoValueTF{##1}
		{
			\IfNoValueTF{##2}
			{#2\prArg##3}
			{#2^{##2}\prArg##3}
		}
		{
			\IfNoValueTF{##2}
			{#2_{##1}\prArg##3}
			{#2_{##1}^{##2}\prArg##3}
		}
	}
}

\DeclareDocumentCommand \fVector {m} {\boldsymbol{#1}}
\DeclareDocumentCommand \fMatrix {m} {\boldsymbol{#1}}

\DeclareDocumentCommand \fFunction {m} {{#1}}
\DeclareDocumentCommand \fSet {m} {\mathcal{#1}}

\DeclareDocumentCommand \fOperator {m} {\mathop{\vphantom{\sum}\mathchoice {\vcenter{\hbox{\Large $#1$}}} {\vcenter{\hbox{\Large $#1$}}}{#1}{#1}}\displaylimits}

\DeclareDocumentCommand \newScalar{r<>m}
{
	\DeclareDocumentCommand #1 {} {{#2}}
}

\DeclareDocumentCommand \newVector{r<>m}
{
	\DeclareDocumentCommand #1 {} {\fVector{#2}}
}

\DeclareDocumentCommand \newMatrix{r<>m}
{
	\DeclareDocumentCommand #1 {} {\fMatrix{#2}}
}

\DeclareDocumentCommand \newProbability{r<>m}
{
	\newProbabilityFormat<#1>{#2}
}

\DeclareDocumentCommand \newFunction{r<>m}
{
	\DeclareDocumentCommand #1 {} {\fFunction{#2}}
}

\DeclareDocumentCommand \newSet{r<>m}
{
	\DeclareDocumentCommand #1 {} {\fSet{#2}}
}

\makeatletter
\def\cleartheorem#1{%
    \expandafter\let\csname#1\endcsname\relax
    \expandafter\let\csname c@#1\endcsname\relax
}
\makeatother

\cleartheorem{corollary}

\Crefname{corollarycount}{Corollary}{Corollaries}

\cleartheorem{definition}

\Crefname{definitioncount}{Definition}{Definitions}

\DeclareDocumentCommand \oK {} {\fOperator{\otimes}}

\DeclareDocumentCommand \reals {} {\mathbb{R}}

\newScalar<\eps>{\varepsilon}

\newVector<\vA>{a}
\newVector<\vB>{b}
\newVector<\vC>{c}
\newVector<\vD>{d}
\newVector<\vE>{e}
\newVector<\vF>{f}
\newVector<\vG>{g}
\newVector<\vH>{h}
\newVector<\vI>{i}
\newVector<\vJ>{j}
\newVector<\vK>{k}
\newVector<\vL>{\ell}
\newVector<\vM>{m}
\newVector<\vN>{n}
\newVector<\vP>{p}
\newVector<\vQ>{q}
\newVector<\vR>{r}
\newVector<\vS>{s}
\newVector<\vT>{t}
\newVector<\vU>{u}
\newVector<\vV>{v}
\newVector<\vW>{w}
\newVector<\vX>{x}
\newVector<\vY>{y}
\newVector<\vZ>{z}
\newVector<\vAlpha>{\alpha}
\newVector<\vBeta>{\beta}
\newVector<\vGamma>{\gamma}
\newVector<\vDelta>{\delta}
\newVector<\vEpsilon>{\varepsilon}
\newVector<\vEps>{\varepsilon}
\newVector<\vZeta>{\zeta}
\newVector<\vEta>{\eta}
\newVector<\vTheta>{\theta}
\newVector<\vTh>{\theta}
\newVector<\vKappa>{\kappa}
\newVector<\vLambda>{\lambda}
\newVector<\vMu>{\mu}
\newVector<\vNu>{\nu}
\newVector<\vXi>{\xi}
\newVector<\vPi>{\pi}
\newVector<\vRho>{\rho}
\newVector<\vSigma>{\sigma}
\newVector<\vTau>{\tau}
\newVector<\vUpsilon>{\upsilon}
\newVector<\vPhi>{\varphi}
\newVector<\vChi>{\chi}
\newVector<\vPsi>{\psi}
\newVector<\vOmega>{\omega}

\newMatrix<\mA>{A}
\newMatrix<\mB>{B}
\newMatrix<\mC>{C}
\newMatrix<\mD>{D}
\newMatrix<\mE>{E}
\newMatrix<\mF>{F}
\newMatrix<\mG>{G}
\newMatrix<\mH>{H}
\newMatrix<\mI>{I}
\newMatrix<\mJ>{J}
\newMatrix<\mK>{K}
\newMatrix<\mL>{L}
\newMatrix<\mM>{M}
\newMatrix<\mN>{N}
\newMatrix<\mP>{P}
\newMatrix<\mQ>{Q}
\newMatrix<\mR>{R}
\newMatrix<\mS>{S}
\newMatrix<\mT>{T}
\newMatrix<\mU>{U}
\newMatrix<\mV>{V}
\newMatrix<\mW>{W}
\newMatrix<\mX>{X}
\newMatrix<\mY>{Y}
\newMatrix<\mZ>{Z}
\newMatrix<\mGamma>{\Gamma}
\newMatrix<\mDelta>{\Delta}
\newMatrix<\mTheta>{\Theta}
\newMatrix<\mLambda>{\Lambda}
\newMatrix<\mXi>{\Xi}
\newMatrix<\mPi>{\Pi}
\newMatrix<\mSigma>{\Sigma}
\newMatrix<\mUpsilon>{\Upsilon}
\newMatrix<\mPhi>{\Phi}
\newMatrix<\mPsi>{\Psi}
\newMatrix<\mOmega>{\Omega}

\newSet<\sA>{A}
\newSet<\sB>{B}
\newSet<\sC>{C}
\newSet<\sD>{D}
\newSet<\sE>{E}
\newSet<\sF>{F}
\newSet<\sG>{G}
\newSet<\sH>{H}
\newSet<\sI>{I}
\newSet<\sJ>{J}
\newSet<\sK>{K}
\newSet<\sL>{L}
\newSet<\sM>{M}
\newSet<\sN>{N}
\newSet<\sP>{P}
\newSet<\sQ>{Q}
\newSet<\sR>{R}
\newSet<\sS>{S}
\newSet<\sT>{T}
\newSet<\sU>{U}
\newSet<\sV>{V}
\newSet<\sW>{W}
\newSet<\sX>{X}
\newSet<\sY>{Y}
\newSet<\sZ>{Z}

\begin{document}

\title{Adaptive n-ary Activation Functions \\ for Probabilistic Boolean Logic}

\author{Jed A. Duersch, Thomas A. Catanach, and Niladri Das%
\thanks{%
Jed Duersch (corresponding author, jaduers@sandia.gov),
Thomas Catanach (tacatan@sandia.gov), and Niladri Das (ndas@sandia.gov)
are with Sandia National Laboratories, Livermore, CA 94550, USA}
\thanks{\copyright 2022 IEEE.
Personal use of this material is permitted. Permission
from IEEE must be obtained for all other uses, in any current or future
media, including reprinting/republishing this material for advertising or
promotional purposes, creating new collective works, for resale or
redistribution to servers or lists, or reuse of any copyrighted
component of this work in other works.
}}

\markboth{Adaptive n-ary Activation Functions}{\copyright 2022 IEEE}

\DeclareDocumentCommand \symN{} {n} 




\maketitle

\thispagestyle{empty}

\begin{abstract}
Balancing model complexity against the information contained in observed data is the central challenge to learning.
In order for complexity-efficient models to exist and be discoverable in high dimensions, 
we require a computational framework that relates a credible notion of complexity to simple parameter representations.
Further, this framework must allow excess complexity to be gradually removed via gradient-based optimization.
Our n-ary, or n-argument, activation functions
fill this gap by approximating belief functions (probabilistic Boolean logic) using logit representations of probability.
Just as Boolean logic determines the truth of a consequent claim from relationships among a set of antecedent propositions,
probabilistic formulations generalize predictions when antecedents, truth tables, and consequents all retain uncertainty.
Our activation functions demonstrate the ability to learn arbitrary logic, such as the binary exclusive disjunction (p xor q) and ternary conditioned disjunction ( c ? p : q ), in a single layer using an activation function of matching or greater arity.
Further, we represent belief tables using a basis that directly associates the number of nonzero parameters to the effective arity of the belief function,
thus capturing a concrete relationship between logical complexity and efficient parameter representations.  
This opens optimization approaches to reduce logical complexity by inducing parameter sparsity.
\end{abstract}




\newScalar<\nArg>{\symN}
\newScalar<\nAry>{\symN}
\newScalar<\nCh>{c}
\newScalar<\wIn>{w_\text{in}}
\newScalar<\wOut>{w_\text{out}}
\newScalar<\wMax>{w_\text{max}}
\newScalar<\wOne>{w_1}
\newScalar<\true>{\textbf{true}}
\newScalar<\false>{\textbf{false}}
\newScalar<\bF>{\boldsymbol{0}}
\newScalar<\bT>{\boldsymbol{1}}
\DeclareDocumentCommand \tF{} {\textbf{false}}
\DeclareDocumentCommand \tT{} {\textbf{true}}


\newVector<\bX>{\xi}
\newVector<\bY>{\psi}
\newVector<\bZ>{\zeta}
\newVector<\bP>{\pi}
\newVector<\vZh>{\hat{z}}


\newMatrix<\mBasis>{C}


\newSet<\sData>{\mathcal{D}}
\newSet<\sTF>{\{\bF,\bT\}}


\newProbability<\pP>{\mathbf{p}}
\newProbability<\pQ>{\mathbf{q}}


\newFunction<\fJ>{J}
\newFunction<\fLen>{\mathcal{\ell}}
\newFunction<\fOrder>{\mathcal{O}}
\newFunction<\fDelta>{\delta}
\newFunction<\fNot>{\mathbf{not}}
\newFunction<\fAnd>{\mathbf{and}}
\newFunction<\fOr>{\mathbf{or}}
\newFunction<\fImp>{\mathbf{imply}}
\newFunction<\fXor>{\mathbf{xor}}
\newFunction<\fXnor>{\mathbf{xnor}}
\newFunction<\fTrue>{\mathbf{true}}
\newFunction<\fFalse>{\mathbf{false}}
\newFunction<\fArgOne>{\mathbf{arg_1}}
\newFunction<\fArgTwo>{\mathbf{arg_2}}
\newFunction<\fNotOne>{\mathbf{not_1}}
\newFunction<\fNotTwo>{\mathbf{not_2}}
\newFunction<\fGen>{f}
\newFunction<\fUnary>{\mathbf{unary}}
\newFunction<\fNary>{\mathbf{n\_ary}}
\newFunction<\sgn>{\mathrm{sign}}
\newFunction<\abs>{\mathrm{abs}}
\newFunction<\fOrAIL>{\mathbf{or}_\mathbf{AIL}}
\newFunction<\fAndAIL>{\mathbf{and}_\mathbf{AIL}}
\newFunction<\fXnorAIL>{\mathbf{xnor}_\mathbf{AIL}}
\newFunction<\fRelu>{\mathbf{relu}}
\newFunction<\fMaxMin>{\mathbf{max\_min}}
\newFunction<\fAma>{\mathbf{ama}}


\DeclareDocumentCommand \entropy{m} {S\!\left[\,#1\,\right]}
\DeclareDocumentCommand \KL{mm} {D\!\left[\,#1\,\middle\|\,#2\,\right]}


\DeclareDocumentCommand \grad{m} {\nabla_{\!#1\,}}

\DeclareDocumentCommand \s{} {\phantom{-}}

\section{Introduction}
\label{sec:introduction}

Theoretically optimal learning, i.e.~assimilating data into models with rigorous mathematical justification for the resulting uncertainty in predictions, may be comprehensively understood within the Bayesian paradigm of reason.
Our investigation into the provenance of prior belief in abstract learning algorithms \cite{Duersch2021} affirms the foundational elegance of minimizing algorithmic complexity, as measured by encoding information \cite{Shannon1948,Kullback1951,Kolmogorov1965}.
This perspective was originally presented by Solomonoff \cite{Solomonoff1960,Solomonoff1964a,Solomonoff1964b,Solomonoff2009} as Algorithmic Probability (AP).
Minimum Description Length (MDL), proposed by Rissanen \cite{Rissanen1983,Rissanen1984}, can be understood as a closely-related optimization approach to identify models that dominate predictions under AP.

Our research into how to suppress model information during training compels the perspective motivating this work:
to acquire complexity-efficient models within a limited computational budget, a learning architecture must
(1) admit a variety of model complexities that have a clear relationship to parameter representations
and (2) differentiable parameter trajectories must connect high-complexity to low-complexity models,
since gradient-based optimization is essential to navigate high-dimensions.

Bayesian or probabilistic formulations of Boolean algebra provide a natural domain to satisfy these requirements.
In general, if we are given $\nAry$~antecedents, propositions with known truth values, then we can determine the truth value of a consequent by referencing a truth table with $2^\nAry$ entries.
When each entry is either true or false, there are $2^{2^\nAry}$ distinct Boolean truth functions.
If we generalize truth functions to \textit{belief functions} and truth tables to \textit{belief tables} \cite{Pearl2014},
then also counting belief tables with entries that are totally uncertain gives $3^{2^\nAry}$ qualitatively distinct belief functions.
Thus, increasing $\nAry$~within a feasible limit allows dimensionality to work in our favor because $\nAry$-ary logic becomes extremely expressive, making it easier to discover useful models.
Extending this to the Bayesian paradigm requires describing not only our belief in plausible antecedent states, but also our belief in the consequent probability conditioned on each distinct antecedent realization.
Rather than employing fixed elementary unary and binary functions ($\fNot$, $\fAnd$, $\fOr$, etc.), wherein each truth table contains only completely certain (true or false) outcomes,
we must allow consequent probability within belief tables to be adjusted during training.
Then, gradient-based optimization makes the full variety of truth functions discoverable.

\subsection{Our Contributions}
\label{sub:our_contributions}

We show how to formulate and efficiently parameterize activation functions that are capable of learning arbitrary $\nAry$-ary probabilistic truth tables, or belief tables,
using logit representations of antecedent probabilities, conditional probabilities, and consequent probabilities.
This is achieved by implementing an activation function that accept $\nAry$-arguments per output element.
As with typical neural networks, the arguments to the activation function are simply computed from matrix multiply on the previous layer.
Each activation output is processed from a parameterized belief table with $2^\nAry$ elements.
For modest arites, $\nAry \leq 6$, such belief tables are feasible to store and compute.
In fact, the number of parameters only scales approximately linearly with $\nAry$ in this domain, since most parameters are contained in the weight matrix that constructs antecedents from the previous layer.
Specifically, for an input width $\wIn$ and output width $\wOut$, the weight matrix contains $\nAry \wOut \wIn$ elements, whereas the belief table contains $\wOut 2^\nAry$. 

Our parameterized activation functions are more flexible than the fixed versions proposed by Lowe \textit{et al.}~\cite{Lowe2021},
which developed logit-space activation functions that explicitly encode approximations of fixed binary operations ($\fAnd$, $\fOr$, and $\fXnor$),
while still maintaining the computational efficiency that results from simple compositions of $\min$, $\max$, and $\abs$ operations.
These compositions also yield well-conditioned gradients for optimization, ensuring that the same gradient magnitudes propagate from consequents to relevant antecedents and belief table elements,
rather than the extreme scaling imbalance that direct probabilistic computations can produce.
Not only does our approach allow a single activation function to learn any binary operation, it also generalizes to learn any $\nAry$-ary logic within a single layer.

\begin{algorithm}[!h]
\caption{Adaptive Unary Activation}
\label{alg:unary}
\fontsize{10}{16}\selectfont
\begin{algorithmic}[1]
\Require $\vY$ is a $\wOut \times 1$ vector of antecedent logits.
\Statex Implied by the shape of $\vY$, $\wOut$ is number of channels.
\Statex $\mA$ is a $\wOut \times 2$ matrix of belief-table logits.
\Ensure $\vZ$ is a $\wOut \times 1$ vector of consequent logits.
\Function{$\vZ=\fUnary$}{$\vY, \mA$}
\State Compute belief-table sums, $\vS$, and differences, $\vD$,
$\begin{bmatrix}
\vS & \vD \\
\end{bmatrix} = \mA \begin{bmatrix}
1 & -1 \\
1 & \s1 \\
\end{bmatrix}$.
\State $\vZ = \frac{1}{2}\left[ \max( \left| \vY \right| + \vS, \left| \vD + \vY \right|) - \max( \left| \vY \right| - \vS,\left| \vD - \vY \right|) \right]$.
\EndFunction
\end{algorithmic}
\end{algorithm}

\begin{algorithm}[!h]
\caption{Adaptive $\nAry$-ary Activation}
\label{alg:nary}
\fontsize{10}{16}\selectfont
\begin{algorithmic}[1]
\Require $\mY$ is a $\wOut \times \nAry$ matrix of antecedent logits.
\Statex Output width $\wOut$ and arity $\nAry$ are implied by $\mY$.
\Statex $\mTheta$ is a $\wOut \times 2^\nAry$ matrix of activation parameters.
\Ensure $\vZ$ is a $\wOut \times 1$ vector of consequent logits.
\Function{$\vZ=\fNary$}{$\mY, \mTheta$}
\Statex \textit{Remark:} Let columns of each matrix be indexed as
\begin{align*}
&\mY = \begin{bmatrix}
\vY_1 & \vY_2 & \cdots & \vY_\nAry
\end{bmatrix}
\quad\text{and} \\
&\mA^{(i)} = \begin{bmatrix}
\vA^{(i)}_1 & \vA^{(i)}_2 & \cdots & \vA^{(i)}_{2^{\nAry-i}}
\end{bmatrix}
\quad\text{for}\quad i \in \{0,1,...,\nAry\}.
\end{align*}
\State Change basis from parameter representations to belief-table logits
($\oK$ is the Kronecker product),
\begin{align*}
\mA^{(0)} = \mTheta \left( \oK_{i=1}^\nAry \begin{bmatrix}
\s1 & 1 \\
-1 & 1 \\
\end{bmatrix} \right).
\end{align*}
\For{$i = 1, 2, \ldots, \nAry$}
\For{$j = 1, 2, \ldots, 2^{\nAry-i}$}
\State $\vA^{(i)}_j = \fUnary\left(\vY_i, \left[ \vA^{(i-1)}_{2j-1}\; \vA^{(i-1)}_{2j} \right]\right)$
\EndFor
\EndFor 
\State Return output, $\vZ = \mA^{(\nAry)}$.
\EndFunction
\end{algorithmic}
\end{algorithm}

Both \Cref{alg:unary} and \Cref{alg:nary} provide simplified forms of these activation functions.
The change of basis in \Cref{alg:nary}, line 2, which we will examine closely in \Cref{sub:sparsity},
allows us to match representation sparsity to the number of antecedents that are actually needed to evaluate the result.
Thus, this basis forges a direct relationship between encoding complexity and logical complexity through sparsity.
Although it is beyond the primary scope of this paper, this association opens optimization approaches to suppress logical complexity by promoting parameter sparsity during training.

We cover background information regarding probabilistic Boolean logic and related work on activation functions in \Cref{sec:background}.
\Cref{sec:derivation} contains our main derivations, analysis relating sparsity to logical complexity,
and a gradient comparison for our approach versus a direct probabilistic computation.
We present experimental results in \Cref{sec:experiments} followed by a brief discussion
of future challenges and concluding remarks in \Cref{sec:discussion}.

\section{Background}
\label{sec:background}

The study of logical relationships between antecedent claims and consequents is the subject of Boolean algebra.
Thus, it may be unsurprising that human biology evolved to replicate elementary Boolean logic operations.
Gideon \textit{et al.}~\cite{Gidon2020} showed that individual human neurons are capable of replicating the behavior of exclusive disjunction, $\fXor$,
a capability that is not possible using a single affine transformation followed by either rectified linear units (ReLU), MaxOut \cite{Goodfellow2013}, or MaxMin \cite{Chernodub2016} activation functions.
Moreover, higher-arity logical operations also capture important relationships that are central to reasoning,
e.g.~the conditioned disjunction \cite{Church1996} is a ternary operation that is so useful for programming, it is written compactly in C as \textit{(~[condition]~?~[result\_if\_true]~:~[result\_if\_false]~)} to capture conditional reasoning.

While deep neural networks have demonstrated astonishing feats of pattern recognition and task optimization,
we remain curious as to whether more efficient architectures exist that build upon these foundations of logic.
The arity, or number of arguments, needed to evaluate a consequent has a clear connection to computational complexity,
and therefore motivates a compelling framework for prior belief in abstract learning algorithms.
Given this, we pursue an approach that is easily capable of learning diverse logical compositions.

\subsection{Remark on Notation}

The notation we adopt builds toward an implementation wherein a network layer will act upon an input vector, $\vX$,
using a linear transformation, $\vY = \mM \vX$, to provide activation function inputs, $\vZ = \fGen(\vY, \mTheta)$,
where $\mTheta$ is an array of activation parameters.
Although we denote the output as $\vZ$ to avoid excessive indexing, it is understood to take on the role of $\vX$ in the next layer, i.e.~$\vX^{(\ell+1)} = \vZ^{(\ell)}$.
To derive the activation function, however, our analysis begins by considering probabilities associated with underlying Boolean random variables,
which we represent using Greek letters $(\bX,\bY,\bZ)$ so that the corresponding logits can be written as the Latin versions $(\vX,\vY,\vZ)$.

\subsection{Probabilistic Boolean Logic}

In Boolean algebra, propositions are held to be either false or true, written as $\bF$ or $\bT$, respectively.
The typical construction introduces basic operations, including the unary $\fNot(\cdot)$ and the binary functions, $\fAnd(\cdot,\cdot)$ and $\fOr(\cdot,\cdot)$.
In this context, \textit{binary} refers to the number of arguments, rather than the values each argument may take.
\textit{Boolean} refers to values themselves, e.g. $\bF$ or $\bT$.
An $\nAry$-ary law of composition, or truth function, accepts $\nAry \geq 1$ arguments to compute an outcome, $\fGen : \sTF^\nAry \mapsto \sTF$.
The arguments are called antecedents, denoted elementwise as $\bY_i \in \sTF$ for $i \in [\nAry]$ or as the tuple $\bY = (\bY_1, \bY_2, \ldots, \bY_{\nAry})$,
and the outcome $\bZ = \fGen(\bY) \in \sTF$ is called the consequent.

These tools of deductive reasoning, however, are only justified in contexts of complete certainty,
the domains of computer science and mathematical proofs that build and analyze consistent axiomatic foundations using formal definitions and their emergent properties.
Bayesian reasoning is an inductive approach, consistent with the laws of probability, to analyze degrees of plausibility among multiple explanations.
Incorporating probability into logical analysis of propositions goes back to the work of Good \cite{Good1950} and De Finetti\cite{DeFinetti1974},
who examined logical combinations of propositions that may be expressed as a composition of negations, conjunctions, and disjunctions,
where antecedents retain some degree of uncertainty.

One begins by constructing a probability space over the set of potential outcomes, i.e.~$\mOmega = \sTF$ for a single claim.
Thus, our belief in a claim, e.g.~that a particular antecedent is true, $\bY_i=\bT$, is represented by the probability mass function $\pP(\bY_i=\bT) \in [0, 1]$, a Bernoulli random variable.
Provided antecedents are independent, we have the usual compositions: negation, $\pP(\lnot\bY_1) = 1-\pP(\bY_1)$; conjunction, $\pP(\fAnd(\bY_1,\bY_2)) = \pP(\bY_1)\pP(\bY_2)$; and disjunction, $\pP(\fOr(\bY_1,\bY_2)) = 1-\pP(\bY_1)\pP(\bY_2)$.
For our purposes, however, we must also incorporate uncertainty into the truth tables themselves.

Pearl \cite{Pearl2014} provides a comprehensive reference for probabilistic reasoning, including generalizing truth tables to belief tables to allow consequent uncertainty that is conditioned upon every possible antecedent realization. 
This general approach forms the joint distribution, including both the consequent and all antecedents, as $\pP(\bZ, \bY_1, \bY_2) = \pP(\bZ \mid \bY_1, \bY_2) \pP(\bY_2, \bY_1)$.
Marginalizing over all mutually exclusive antecedent states gives the general binary belief function,
\small
\begin{align*}
\pP(\bZ) &= \pP(\bZ \mid \bY_1, \bY_2) \pP(\bY_1, \bY_2) +
\pP(\bZ \mid \lnot\bY_1, \lnot\bY_2) \pP(\lnot\bY_1, \lnot\bY_2)\\
&+ \pP(\bZ \mid \lnot\bY_1, \bY_2) \pP(\lnot\bY_1, \bY_2) + \pP(\bZ \mid \bY_1, \lnot\bY_2) \pP(\bY_1, \lnot\bY_2).
\end{align*}
\normalsize

We can form any higher-arity belief function by specifying the desired conditional probabilities and evaluating it with our belief in the antecedent distribution.
When every conditional probability is certain, we recover a logical truth function.
For example, exclusive disjunction, $\fXor(\cdot,\cdot)$, can be obtained by setting $\pP(\bZ \mid \bY_1, \lnot\bY_2)=\pP(\bZ \mid \lnot\bY_1, \bY_2)=1$ and $\pP(\bZ \mid \bY_1, \bY_2) =\pP(\bZ \mid \lnot\bY_1, \lnot\bY_2) =0$.
Substitution with antecedent independence gives $\pP(\fXor(\bY_1,\bY_2)) = \pP(\bY_1)(1-\pP(\bY_2)) + (1-\pP(\bY_1))\pP(\bY_2)$.
This is equivalent to computing the expected consequent probability using our belief in antecedents as the view of expectation, which bears an important connection to how we understand rational measures of information \cite{Duersch2020}.
When we have a set of $\nAry$ antecedents, a complete belief table must contain consequent probabilities for every possible antecedent realization, the $\nAry$-ary Cartesian product $\{\bT, \bF\}^\nAry$ or vertices of a hypercube in $\nAry$ dimensions.

\subsection{Approximating Logical Uncertainty}

The concept of fuzzy logic was introduced and developed in the seminal work of Zadeh \cite{Zadeh1975,Zadeh1988}, building on his concept of fuzzy sets \cite{Zadeh1965},
wherein elements of a set are present on a continuum of \textit{grades of membership}.
By building off of the set-theoretic concepts of union and intersection, respectively formulated for grades of membership using $\max$ and $\min$ operations,
Zadeh obtains fuzzy logic versions of the corresponding binary operations ($\fOr$ and $\fAnd$, respectively).
Thus, MaxMin activation functions can be understood to approximate logical uncertainty associated with fixed complementary pairs of binary operations.

\begin{figure}[h!]
	\centering
	\includegraphics[width=0.49\textwidth]{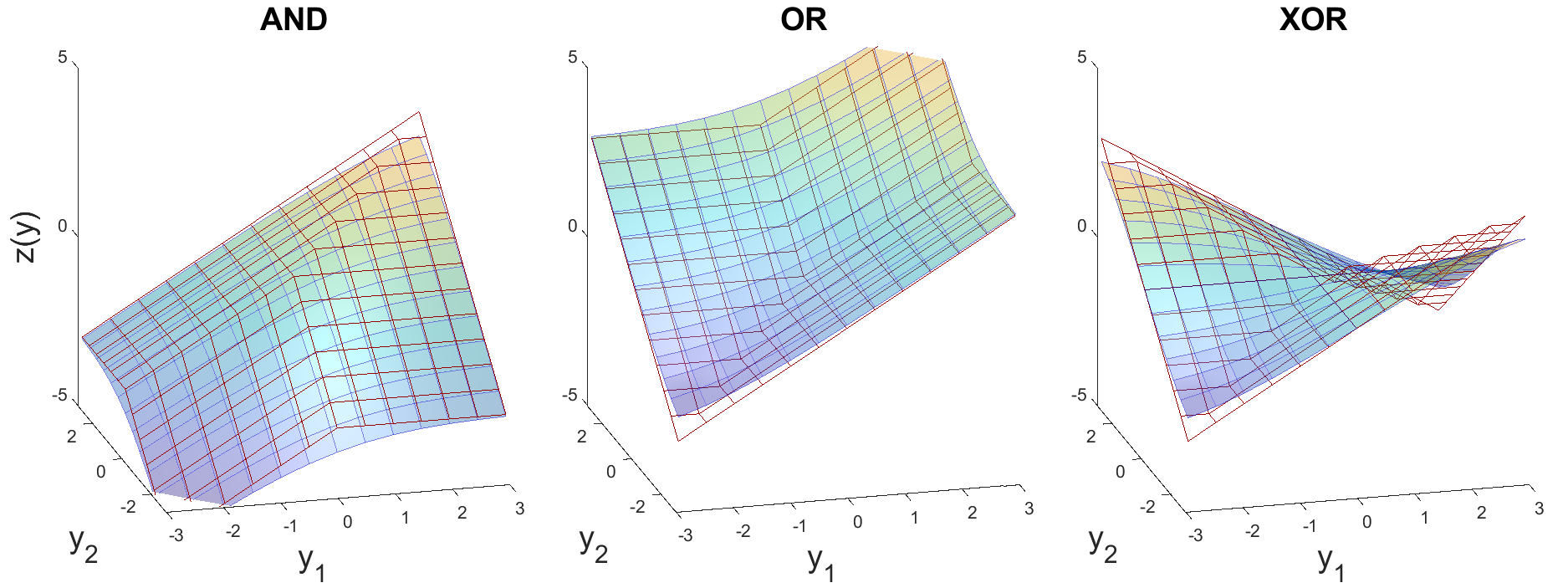}
	\caption{
	Comparison of exact logit-space binary operations (surfaces) with AIL formulations (red grids).
	}
  \label{fig:binary_vis}
\end{figure}

Our approach is somewhat similar to fuzzy logic,
but we derive our activation functions systematically from Log-Sum-Exponential-Max (LSEM) approximations to logit representations of antecedents, truth-table probabilities, and consequents.
See \Cref{sub:lsem} for details.
We observe that LSEM approximations can be applied to any binary truth function and, with modest algebraic manipulation, the Approximate Independent Logit (AIL) formulations presented by Lowe \cite{Lowe2021} are recovered,
\begin{itemize}
\item $\fAndAIL(\mY) = \min(\vY_1, \vY_2, \vY_1 + \vY_2)$
\item $\fOrAIL(\mY) = \max(\vY_1, \vY_2, \vY_1 + \vY_2)$
\item $\fXnorAIL(\mY) = \sgn(\vY_1 \vY_2)\min(\abs(\vY_1),\abs(\vY_2))$
\end{itemize}
\Cref{fig:binary_vis} visualizes these approximations for comparison to the exact logit-space equivalents.
These are not, however, the functions we use in practice.
By parameterizing belief tables, our approach becomes general, allowing consequent uncertainty to be adjusted within higher-arity belief functions as merited by the training data.
 

\subsection{Other Logic-Related Work}

Payani and Fekri \cite{Payani2019} propose a novel learning paradigm for deep neural networks using differentiable binary operators, $\fAnd$, $\fOr$, and $\fXor$.
They show how these elementary operators combine in simple and meaningful ways to form Neural Logic Networks (NLNs)
that outperform state-of-the-art neural inductive logic programming (ILP) for benchmark tasks that include addition, multiplication, and sorting.

Binarized neural networks learn models with equivalent Boolean logic representations that can be formally analyzed with logic-based tools, such as SAT solvers.
Narodytska \textit{et al.}~\cite{Narodytska2020} analyze the efficiency of architectural choices for these networks and discuss how they affect the performance of logic-based reasoning.
They also propose training procedures to obtain simpler network for SAT solvers that avoid sacrificing accuracy on a primary task.

Tavares \textit{et al.}~\cite{Tavares2020} show that deep learning can incorporate several families of Boolean functions, and relatively small and shallow neural networks find good approximations of these functions.
They also discuss the difficulty of learning complex Boolean formulas in Conjunctive Normal Form (CNF), i.e.~as a conjunction of several simple clauses.
Their analysis shows that this difficulty arises because, as the number of clauses increases, fewer input states satisfy all clauses simultaneously to yield a positive example of the composition.
Since our approach approximates marginalization over an entire belief table, every example isolates relevant belief table parameters and informs model updates.
We compare the effects of backpropagation using a direct probabilistic marginalization versus our approach more closely in \Cref{sub:backprop}.

\section{Derivation of $\nAry$-ary Activation Functions}
\label{sec:derivation}

Our objective is to transform an input width of $\wIn$ neurons, representing plausible truth states for a set of propositions, to $\wOut$ outputs.
To emphasize that our representation of plausible inputs states is just an approximation, and that it depends on the particular inputs to a network,
we write input distributions as $\pQ(\bX_j)$ for $j \in [\wIn]$, rather than $\pP(\bX_j)$.
This also avoids confusion with prior belief or needing to include extra conditional variables to provide the distinction.
The probability that $\bX_j=\bT$ is simply $\pQ(\bX_j)$ and the probability that $\bX_j=\bF$ is $\pQ(\lnot\bX_j) = 1 - \pQ(\bX_j)$.

Since the activation function will mimic $\nAry$-ary belief functions for each output channel, the leading linear transformation must isolate $\wOut\times\nAry$ antecedents.
If we were operating on probabilities directly, we would have to constrain these transformations to convex combinations of inputs,
i.e.~all matrix entries would need to be nonnegative with rows summing to 1, to ensure that the results represent coherent probabilities.
Working in logit space, however, easily avoids this complication,
\begin{displaymath}
\vX_j = \log\!\left(\frac{\pQ(\bX_j)}{\pQ(\lnot\bX_j)}\right),
\quad\text{so that}\quad
\pQ(\bX_j) = \frac{1}{1+\exp(-\vX_j)}.
\end{displaymath}
Then, any linear transformation $\vY = \mM \vX$ with $\mM \in \mathbb{R}^{\wOut \nAry \times \wIn}$ will provide antecedents with coherent probabilistic interpretations.
The matricization, $\mY = \mathrm{matrix}(\vY, \wOut \times \nAry)$,
allows us to identify the consequent index, $k \in [\wOut]$, with the row and the corresponding argument index, $i \in [\nAry]$, with the column.
In principle, the antecedent probabilities become $\pQ(\bY_{ki}) = [1+\exp(-\vY_{ki})]^{-1}$, although these are never actually computed.
Likewise, each consequent channel, $\vZ_k$, also has a probability implied by the corresponding logit, $\pQ(\bZ_{k}) = [1+\exp(-\vZ_{k})]^{-1}$.

The belief table that defines the activation function must articulate the probability of a consequent state $\bZ_k=\bT$ for each realization of antecedents,
i.e.~storing $\pP(\bZ_k \mid \bY_k)$ where the tuple $\bY_k = (\bY_{k1}, \bY_{k2}, \ldots, \bY_{k\nAry})$ varies over all values in $\sTF^\nAry$.
In order to build toward \nAry-ary activations, we begin with the unary case, $\nAry=1$.
To propagate our belief from $\pQ(\bY_{k1})$ to $\pQ(\bZ_k)$, we have the elementary marginalization,
\begin{align}
\label{eqn:unary_marginalization}
\pQ(\bZ_k) = \pP(\bZ_k \mid \lnot\bY_{k1}) \pQ(\lnot\bY_{k1}) + \pP(\bZ_k \mid \bY_{k1}) \pQ(\bY_{k1}). 
\end{align}

Just as we operate on logit representations of antecedents and consequents, we also use logits to represent belief-table entries.
In the unary case, the belief table is a matrix $\mA \in \mathbb{R}^{\wOut \times 2}$ so that the two elements in row $k$, $\vA_{k0}$ and $\vA_{k1}$
\footnote{We typically use 1-based indexing, but for this work it is easier to see the relationship between the column index and the antecedent state
if count from zero in binary.}, indicate
\small
\begin{align}
\label{eqn:unary_belief_table_logits}
\pP(\bZ_k \mid \lnot\bY_{k1}) = \frac{1}{1+e^{-\vA_{k0}}},
\quad
\pP(\bZ_k \mid \bY_{k1}) = \frac{1}{1+e^{-\vA_{k1}}}.
\end{align}
\normalsize
Going forward, we will call elements $\vA_{k j}$ of the belief table $\mA$ the belief table logits.
Combining \Cref{eqn:unary_marginalization,eqn:unary_belief_table_logits}, using logit representations of both antecedents and consequents, gives the general unary logit-space marginalization,
\begin{align}
\label{eqn:unary_logit_marginalization}
\vZ_k
&=\log\!\left( e^{\frac{\vA_{k0} - \vA_{k1}  - \vY_{k1}}{2}} + e^{\frac{\vA_{k0} + \vA_{k1}  - \vY_{k1}}{2}} + e^{\frac{-\vA_{k0} + \vA_{k1}  + \vY_{k1}}{2}} \right. \nonumber \\
&\left. + e^{\frac{\vA_{k0} + \vA_{k1}  + \vY_{k1}}{2}} \right) -\log\!\left( e^{\frac{-\vA_{k0} + \vA_{k1}  - \vY_{k1}}{2}} + e^{\frac{-\vA_{k0} - \vA_{k1}  - \vY_{k1}}{2}} \right. \nonumber \\
&\left. + e^{\frac{-\vA_{k0} - \vA_{k1}  + \vY_{k1}}{2}} + e^{\frac{\vA_{k0} - \vA_{k1}  + \vY_{k1}}{2}} \right).
\end{align}

\Cref{eqn:unary_logit_marginalization} shows that even this unary case is too complicated to be used directly, and would defeat the convenience of logit-space activation functions. 
Both $\log$ and $\exp$ are also relatively expensive operations that require numerical library implementations;
they are not implemented as single assembly operations, whereas the alternatives we will use ($\max$, $\min$, and $\abs$) often are.
Implementing such activation functions in a large neural network would become prohibitive if we had to repeatedly use many expensive operations.

\subsection{Log-Sum-Exponential Max Approximation}
\label{sub:lsem}

Fortunately, these constructions allow a simple and efficient approximation, the Log-Sum-Exponential Max (LSEM)
\begin{align*}
\log\!\left( \sum_{i=1}^n \exp( \vX_i ) \right) \approx \max_{i=1}^n \vX_i.
\end{align*}
The approximation error is bound by $0 \leq \log\!\left( \sum_{i=1}^n \exp( \vX_i ) \right) - \max_{i=1}^n \vX_i \leq \log(n)$
and we can apply maximization in steps, using only a couple terms at a time, without introducing any compounded approximation errors.
When we apply LSEM approximation to \Cref{eqn:unary_logit_marginalization}, and simplify results with the identity $\max(a, b) = (a + b + |a - b|)/2$, we have
\begin{align}
\label{eqn:unary_lsem}
\vZ_k = \frac{1}{2}&\left[ \max( |\vY_{k1}| + \vS_k, |\vD_k + \vY_{k1}| )\right. \nonumber \\
&\left. - \max( |\vY_{k1}| - \vS_k, |\vD_k - \vY_{k1}| ) \right],
\end{align}
where $\vS_k = \vA_{k0} + \vA_{k1}$ and $\vD_k = \vA_{k1} - \vA_{k0}$.
For obvious reasons, we call $\vS$ the sum and $\vD$ the difference.
\Cref{alg:unary} shows the basic implementation.

\begin{figure}[h!]
	\centering
	\includegraphics[width=0.49\textwidth]{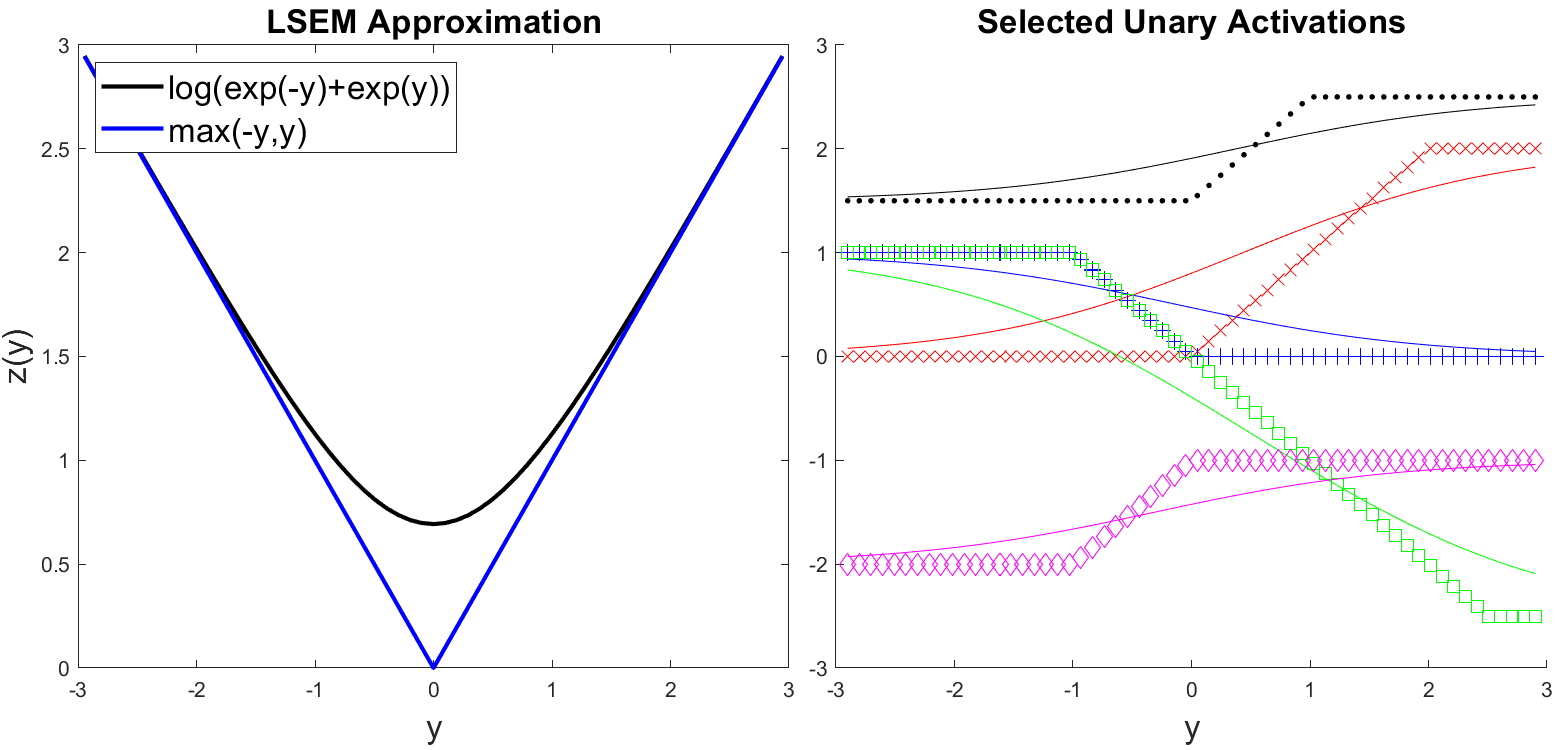}
	\caption{
	\textit{Left:} Illustration of LSEM approximation error.
	Note that the LSEM gradient is equal to the slope of the secant joining the current input-output pairing to the asymptotic extreme.
	\textit{Right:} Illustration of select unary activation functions utilizing the LSEM approximation.
	In each case, the belief table specifies the asymptotic logit values on the left and the right.
	Our activation functions reach the asymptotic limits early.
	Otherwise, for uncertain inputs with a logit near zero, the LSEM increases uncertainty by hewing to an outcome closer to zero.
	The second function down from the right (in red) mimics a rectified linear unit, but with confidence capped at the maximum logit value in the belief table.
	This formulation allows a wider variety of elementwise activation functions to be adaptively determined. 
	}
  \label{fig:visUnary}
\end{figure}

\Cref{fig:visUnary} illustrates the LSEM approximation acting on an antisymmetric pair of logits (left) as well as several resulting unary activation functions (right), shown with marker symbols,
and compares them to the exact marginalization logits (solid lines).
We remark that ReLU activation corresponds to a unary LSEM approximation wherein output symmetry of belief is only broken when the antecedent has a positive logit.

\subsection{Computing $\nAry$-ary Activation Functions}

Using the same notation as the unary case, each consequent index $k \in [\wOut]$ has a tuple of antecedents $\bY_k = (\bY_{k1}, \bY_{k2}, \ldots, \bY_{k\nAry})$.
If every antecedent probability is independent, i.e.~$\pQ(\bY_k) = \prod_{i=1}^\nAry \pQ(\bY_{ki})$,
then it is possible to write the marginalization of any $\nAry$-ary belief table as a nested formula,
\small
\begin{align}
\label{eqn:nested_marginalization}
\pQ(\bZ_k) = \sum_{\bY_{k\nAry}} \pQ(\bY_{k\nAry}) \cdots \sum_{\bY_{k2}}\pQ(\bY_{k2}) \sum_{\bY_{k1}=0}^1 \pQ(\bY_{k1}) \pP(\vZ_k \mid \bY_k).
\end{align}
\normalsize
This will provide a simple iterative method to compute higher-arity activation functions by removing one dimension from the effective belief table at a time.
To express the formulas that follow compactly, we represent a sub-tuple of antecedents as $(\bY_k)_{i}^{\nAry} = (\bY_{ki}, \bY_{k(i+1)}, \ldots, \bY_{k\nAry})$.
These formulas are understood to hold for every realization, $(\bY_k)_{i}^{\nAry} \in \sTF^{\nAry-i+1}$.

Starting with the first antecedent marginalization, we have
\begin{align}
\label{eqn:nary_first_marginalization}
\pQ(\bZ_k \mid (\bY_k)_2^\nAry) =& \pP(\bZ_k \mid \lnot\bY_{k1}, (\bY_k)_2^\nAry) \pQ(\lnot\bY_{k1}) \nonumber \\
&+ \pP(\bZ_k \mid \bY_{k1}, (\bY_k)_2^\nAry) \pQ(\bY_{k1}).
\end{align}
As with the unary case, the conditional probabilities that form the belief tables are represented as logits in a matrix, $\mA^{(0)} \in \mathbb{R}^{\wOut \times 2^\nAry}$,
with the superscript $(0)$ serving a purpose that will become clear shortly.
A single column index implicitly tracks all possible antecedent states by mapping the binary representation, counting from zero, to each antecedent realization.
Starting with all false, this stores flips in the first antecedent within consecutive columns, flips in the second antecedent in consecutive 2-blocks, and so on.
Subsequent marginalizations are
\begin{align}
\label{eqn:nary_subseq_marginalizations}
\pQ(\bZ_k \mid (\bY_k)_{i+1}^\nAry) =&  \pQ(\bZ_k \mid \lnot\bY_{ki}, (\bY_k)_{i+1}^\nAry) \pQ(\lnot\bY_{ki}) \nonumber \\
&+ \pQ(\bZ_k \mid \bY_{ki}, (\bY_k)_{i+1}^\nAry) \pQ(\bY_{ki}),
\end{align}
where $i = 2, \cdots, \nAry$.
When $i = \nAry$, we interpret $(\bY_k)_{\nAry+1}^\nAry$ as the empty tuple, i.e.~$\pQ(\bZ_k \mid (\bY_k)_{\nAry+1}^\nAry) = \pQ(\bZ_k)$.

Since the conditional probabilities we compose in each subsequent marginalization serve the same role as a lower-arity belief table,
we can also represent them as matrices of logits, $\mA^{(i)} \in \mathbb{R}^{\wOut \times 2^{\nAry - i}}$,
where the superscript $(i)$ indicates the number of antecedents that have been marginalized thus far.
Specifically, for $i \in [\nAry]$, we have
\small
\begin{align*}
\vA_{k j}^{(i)} = \log\!\left(\frac{\pQ( \bZ_k \mid (\bY_k)_{i+1}^\nAry )}{\pQ( \lnot\bZ_k \mid (\bY_k)_{i+1}^\nAry )} \right),
\,\,\text{with}\,\,
j = \textrm{index}\!\left( (\bY_k)_{i+1}^\nAry \right).
\end{align*}
\normalsize
Thus, $\mA^{(0)}$ indicates that the initial belief tables have not marginalized any antecedents.
\Cref{alg:nary} shows how \Cref{eqn:nary_first_marginalization,eqn:nary_subseq_marginalizations} can be implemented easily.
If we have $\nAry$ arguments, the first marginalization acts on $2^{\nAry-1}$ edges of a hypercube,
the second acts on $2^{\nAry-2}$ edges of the previous result, and so forth until the last argument marginalizes over a single edge.

While the total number of unary marginalizations, $\wOut (2^{\nAry}-1)$, may seem prohibitive,
this computation is reasonable for modest values of $\nAry$.
Although our logic experiments in \Cref{sub:logic-mlp} only test up to $\nAry = 6$,
during this research we successfully tested transformations using $\wIn = \wOut = 32$ and $\nAry = 8$ without a significant computational burden.
At these specifications, the number of parameters in the preceding matrix $M$ is exactly equal to the number of parameters needed to construct the belief table, $\mA^{(0)}$.
In practice, however, we anticipate $\nAry=4$ being both reasonably expressive and computationally efficient.

\subsection{Basis for Sparse Representations of Effective Arity}
\label{sub:sparsity}

We can construct the belief-table logits with a change of basis that allows the number of stored nonzero parameters to reflect the number of arguments that are actually needed to evaluate the function.
This allows us to prefer lower-arity logic during training by optimization methods that increase parameter sparsity.
For an $\nAry$-ary activation function with $\wOut$ channels, we store a parameter matrix, $\mTheta \in \mathbb{R}^{\wOut \times 2^{\nAry}}$,
and form a basis $\mB$ from Kronecker products to obtain the belief table matrix,
\begin{align}
\label{eqn:change_of_basis}
\mA^{(0)} = \mTheta \mB
\quad\text{where}\quad
\mB = \oK_{i=1}^\nAry \begin{bmatrix}
\s1 & 1 \\
-1 & 1 \\
\end{bmatrix}.
\end{align}

The $i$\textsuperscript{th} factor in the Kronecker product indicates how corresponding columns in $\mTheta$ will affect the $i$\textsuperscript{th} dimension of a belief table, viewed as a hypercube.
This effectively separates parameters into those that have a symmetric impact, i.e.~those that multiply the first row of factor $i$, and those that have an antisymmetric impact, multiplying the second.
It is not an accident that the matrix in \Cref{alg:unary} that generates the sum and difference vectors, symmetric and antisymmetric terms, is exactly twice the inverse of one of these factors.
By setting all parameters that have an antisymmetric impact on a given antecedent to zero, we create a belief function that is indifferent to the state of that antecedent and, thus, has lower effective arity.
The following theorem formalizes this claim. The proof is contained in \Cref{sec:appendix}.

\subsubsection{Theorem: Irrelevant Antecedents and Parameter Zeros}
Let $\pP(\bZ_k \mid \bY_k)$ represent the $k$\textsuperscript{th} $\nAry$-ary belief table.
Evaluating the tuple of antecedents, $\bY_k = (\bY_{k1}, \bY_{k2}, \ldots, \bY_{k\nAry})$, at any vertex of the hypercube, $\bY_k \in \sTF^\nAry$,
gives the probability of the consequent $\bZ_k$.
Let $\vA_k \in \reals^{2^\nAry}$ be the row vector of logit representations, i.e.~$\vA_{kj} = \text{logit}( \pP(\bZ_k \mid \bY_k) )$ where the column index $j$ maps to $\bY_k = \text{bits}(j)$, counting from zero.
The subset of irrelevant antecedents is defined as
\begin{align*}
\mathcal{I} = &\left\{ \bY_{ki} \mid \pP(\bZ_k \mid \bY_k) = \pP(\bZ_k \mid (\bY_{k1}, \ldots, \lnot\bY_{ki}, \ldots, \bY_{k\nAry})) \right. \\
&\left. \text{for all}\quad \bY_k \in \sTF^\nAry \right\}.
\end{align*}
Using the change of basis, \Cref{eqn:change_of_basis}, if we have $\bY_{ki} \in \mathcal{I}$ then $\vTheta_{kj} = 0$ for all $\text{bit}_i(j)=1$.

\subsubsection{Change of Basis Examples}
A single soft $\fAnd$ operation, using logit magnitudes $\alpha$ in the belief table, can be represented with a binary activation function as
$\mA^{(0)} = \alpha
\begin{bmatrix}
-1 & -1 & -1 & 1 \\
\end{bmatrix}$.
The change of basis gives parameters $\mTheta = \alpha\begin{bmatrix}
\frac{-1}{2} & \frac{1}{2} & \frac{1}{2} & \frac{1}{2} \\
\end{bmatrix}$.
If we represent the same soft $\fAnd$ as a ternary activation function on arguments 1 and 3, we obtain the same nonzeros, but interspersed among zeros for all antisymmetric parameters in argument 2, i.e.~when $\text{bit}_2(j)=1$ or $j = \{2,3,6,7\}$, 
$\vTheta =
\alpha
\begin{bmatrix}
-\frac{1}{2} & \frac{1}{2} & 0 & 0 & \frac{1}{2} & \frac{1}{2} & 0 & 0\\
\end{bmatrix}$.

An immediate consequence of this theorem is that the number of nonzero parameters, $n_{nz}$, is bound by the effective arity, $m = \nAry - |\mathcal{I}|$, as $n_{nz} \leq 2^m$.
Curiously, if we extend this reasoning to derive the complexity of any belief function in this basis, we find that some functions are simpler than we might expect.
\Cref{fig:soft_binary} shows several binary activation functions.

\begin{table}[h]
\centering
\caption{Representations for Selected Binary Activation Functions}
\begin{tabular}{| r || r | r | r | r || r | r | r | r || r |} \hline
\hspace{-1mm}Operation\hspace{-1mm} & \hspace{-1mm}$\vA_{00}$\hspace{-1mm} & \hspace{-1mm}$\vA_{01}$\hspace{-1mm} & \hspace{-1mm}$\vA_{10}$\hspace{-1mm} & \hspace{-1mm}$\vA_{11}$\hspace{-1mm} & \hspace{-1mm}$\vTh_{1}$\hspace{-1mm} & \hspace{-1mm}$\vTh_{2}$\hspace{-1mm} & \hspace{-1mm}$\vTh_{3}$\hspace{-1mm} & \hspace{-1mm}$\vTh_{4}$\hspace{-1mm} \\ \hline
$\mathbf{true}$ & $1$ & $1$ & $1$ & $1$ & $1$ & $0$ & $0$ & $0$ \\ \hline
$\mathbf{arg_1}$ & $-1$ & $1$ & $-1$ & $1$ & $0$ & $1$ & $0$ & $0$ \\ \hline
$\mathbf{not_2}$ & $1$ & $1$ & $-1$ & $-1$ & $0$ & $0$ & $-1$ & $0$ \\ \hline
$\mathbf{xor}$ & $-1$ & $1$ & $1$ & $-1$ & $0$ & $0$ & $0$ & $-1$ \\ \hline
$\mathbf{relu_1}$ & $0$ & $1$ & $0$ & $1$ & $\nicefrac{1}{2}$ & $\nicefrac{1}{2}$ & $0$ & $0$ \\ \hline
$\mathbf{relu_{\lnot2}}$ & $1$ & $1$ & $0$ & $0$ & $\nicefrac{1}{2}$ & $0$ & \hspace{-1mm}$\nicefrac{-1}{2}$ & $0$ \\ \hline
$\mathbf{relu_{xor}}$ & $0$ & $1$ & $1$ & $0$ & $\nicefrac{1}{2}$ & $0$ & $0$ & \hspace{-1mm}$\nicefrac{-1}{2}$ \\ \hline
$\mathbf{imply}$ & $1$ & $-1$ & $1$ & $1$ & $\nicefrac{1}{2}$ & \hspace{-1mm}$\nicefrac{-1}{2}$ & $\nicefrac{1}{2}$ & $\nicefrac{1}{2}$ \\ \hline
$\mathbf{imply^*}$ & $0$ & $-1$ & $0$ & $1$ & $0$ & $0$ & $\nicefrac{1}{2}$ & $\nicefrac{1}{2}$ \\ \hline
$\mathbf{and}$ & $-1$ & $-1$ & $-1$ & $1$ & \hspace{-1mm}$\nicefrac{-1}{2}$ & $\nicefrac{1}{2}$ & $\nicefrac{1}{2}$ & $\nicefrac{1}{2}$ \\ \hline
$\mathbf{or}$ & $-1$ & $1$ & $1$ & $1$ & $\nicefrac{1}{2}$ & $\nicefrac{1}{2}$ & $\nicefrac{1}{2}$ & \hspace{-1mm}$\nicefrac{-1}{2}$ \\ \hline
$\mathbf{and^*}$ & $-1$ & $0$ & $0$ & $1$ & $0$ & $\nicefrac{1}{2}$ & $\nicefrac{1}{2}$ & $0$ \\ \hline
\end{tabular}
\label{fig:soft_binary}
\end{table}


Note the difference between the canonical formulation of material implication, $\fImp$ or $p \rightarrow q$, and the formulation employing a stronger notion of uncertainty, $\fImp^*$.
For $\fImp$, the outcome is $q$ if the premise $p$ is true, otherwise the outcome is true.
In contrast, the more reserved $\fImp^*$ remains uncertain when the premise is false, and sustains a reduction in representation complexity as a result.
Another hybrid function, $\mathbf{and^*}$, also reduces representation complexity in comparison to both $\fAnd$ and $\fOr$ by only reporting agreement between both arguments, otherwise remaining uncertain.

\subsection{Gradient Comparison}
\label{sub:backprop}

The gradient structure of an activation function can have a critical impact on practical trainability.
For an activation function to be used in modern deep neural networks, it must avoid the problem of vanishing gradients.
When we examine the gradient structure of the unary LSEM in \Cref{alg:unary}, i.e.~\Cref{eqn:unary_lsem}, 
we note that each maximization argument is simply a signed sum of an antecedent logit and two belief table logits.
This means that when we subtract both maxima, the signs either cancel or double, and then the result is divided by two.
Thus, $\vZ_k$ is effectively a sparse signed sum over $\vY_{k1}$, $\vA_{k0}$, and $\vA_{k1}$,
i.e.~each term is multiplied by either $-1$, $0$, or $1$.
Given loss $J$, backpropagation easily gives gradient magnitudes
\begin{align*}
\left| \frac{\partial J}{\partial \vY_{k1} } \right|,\, \left| \frac{\partial J}{\partial \vA_{k0} } \right|, \,\text{and}\, \left| \frac{\partial J}{\partial \vA_{k1} } \right|
\in \left\{0, \left| \frac{\partial J}{\partial \vZ_{k} } \right| \right\}.
\end{align*}

Further, if $\frac{\partial J}{\partial \vA_{k0}} \neq 0$ then $\frac{\partial J}{\partial \vA_{k1}} = 0$ and vice versa.
To see this, note that if $|y| + s \geq |d + y|$ then we must have $|y| - s = 2|y| - \left(|y| + s\right) \leq 2|y| - |d + y| \leq |d - y|$,
causing sign cancellation to occur for either $\vA_{k0}$ or $\vA_{k1}$, since both an $s$ and $d$ term survive.
Similarly, mapping $(s,y)\mapsto(-s,-y)$ shows that if $|y| - s \geq |d - y|$ then $|y| + s \leq |d + y|$, also causing one cancellation.
It is also possible for both $|d + y| > |y| + s$ and $|d - y| > |y| - s$.
Let us define signs, $\sigma_{+}=\pm 1$ and $\sigma_{-}=\pm 1$, so that $|d+y| = \sigma_{+}(d+y)$ and $|d-y| = \sigma_{-}(d-y)$.
Then $|d+y| + |d-y| = (\sigma_{+} + \sigma_{-})d + (\sigma_{+} - \sigma_{-})y > 2|y|$.
This only holds if the signs are equal, $\sigma_{+} = \sigma_{-}$, and $|d| > |y|$,
thus causing $d$ dependence to vanish from the difference so that both $\frac{\partial J}{\partial \vA_{k0}} = \frac{\partial J}{\partial \vA_{k1}} = 0$.

Since \Cref{alg:nary} is a composition of several calls to \Cref{alg:unary},
each intermediate result may only have, at most, one nonzero derivative with respect to a preceding partially-marginalized belief table.
That is, if $\frac{\partial \vA^{(i)_{k\,j}}}{\partial \vA^{(i-1)}_{k\,2j-1}} = \pm 1$ then $\frac{\partial \vA^{(i)_{k\,j}}}{\partial \vA^{(i-1)}_{k\,2j}} = 0$ and vice versa.
Thus, the magnitude of the loss gradients is constant or zero through the backpropagation sequence to all antecedents and the gradient with respect to belief table elements contains, at most, a single nonzero.
It easily follows that the change of basis, \Cref{eqn:change_of_basis}, transmits the same gradient magnitude to all stored parameters for the belief table. 

In contrast, if we directly marginalized probabilities in \Cref{eqn:nary_first_marginalization,eqn:nary_subseq_marginalizations},
then backpropagation on the last marginalization would give
\small
\begin{align*}
&\frac{\partial J}{\partial \pQ(\bZ_k \mid \bY_{kn}) } = \frac{\partial J}{\partial \pQ(\bZ_k)} \pQ(\bY_{kn}), \\
&\frac{\partial J}{\partial \pQ(\bZ_k \mid \lnot\bY_{kn}) } = \frac{\partial J}{\partial \pQ(\bZ_k)}\left(1- \pQ(\bY_{kn})\right), \quad\text{and}\\
&\frac{\partial J}{\partial \pQ(\bY_{kn}) } = \frac{\partial J}{\partial \pQ(\bZ_k) }\left( \pQ(\bZ_k \mid \bY_{kn}) - \pQ(\bZ_k \mid \lnot\bY_{kn}) \right).
\end{align*}
\normalsize
Since the marginalizations can be computed in any order, analogous formulas hold for any $\bY_{ki}$ for $i \in [\nAry]$.
The point is the magnitude of each gradient is attenuated by each marginalization, either multiplying by a probability ($\in [0, 1]$) or by a difference of probabilities ($\in [-1, 1]$),
and potentially creating a vanishing gradient problem.
If we composed several belief functions in this form, the gradient with respect to initial belief tables could be orders of magnitude smaller than with respect to final belief tables.
In contrast, our activation functions eliminate this issue by ensuring that equivalent gradient magnitudes are transmitted through the activation function.

\section{Numerical Experiments}
\label{sec:experiments}

Our primary objective is to forge a path toward logical complexity suppression in support of principled uncertainty quantification in automatic learning algorithms.
While our activation functions promote that goal, reaching it will still require additional advances that are beyond the scope of this contribution.
Although our activation functions may reduce the number of layers needed to express logical relationships between latent antecedents and consequents,
the number of parameters within fixed layers increases, approximately linearly with $\nAry$ for $\nAry \leq 6$, to retain the same input and output widths.
Without having new training approaches that efficiently induce sparsity and thereby reduce the number of effective parameters,
a fair comparison to other activation functions is both difficult and somewhat premature.

These experiments simply demonstrate that our activation functions perform as intended and are already competitive with other activation functions on standard networks and learning problems.
The first set of numerical experiments examines the ability of a variety of activation functions to recover truth functions with a range of arities
and shows that our activation functions do indeed learn high-arity logic efficiently.
We then compare activation functions on a simple Convolutional Neural Network (CNN) for MNIST \cite{LeCun2010} as well as a standard CNN for CIFAR-10 \cite{Krizhevsky2009}.
Since we do not yet have the tools to efficiently constrain the number of effective degrees of freedom in parameters,
we frame CNN comparisons in terms of layers with equal output widths, i.e.~all activation functions compute the same number of output neurons in respective networks.
This probes how efficiently each neuron transfers relevant information.
\Cref{sec:appendix} contains a table for each experiment with the number of parameters used in each architecture.

\subsection{Learning Higher-Arity Logic}
\label{sub:logic-mlp}
This set of experiment is designed to investigate how easily various activation functions are able to learn random truth functions.
Each ground truth consists of 32 independent input variables and 32 output variables and a fixed ground truth-function arity $\gamma$.
Each output function is generated by selecting a random subset of $\gamma$ distinct inputs, providing antecedents,
and a random truth function is then generated by drawing $2^\gamma$ truth-table entries selected from $\sTF$ uniformly.
With these mappings, we can generate data sets by drawing 32 independent inputs and applying the 32 corresponding truth functions per training case.
True and false values are replaced with maximum and minimum logits, $\pm6.91$, corresponding to probabilities $0.999$ and $0.001$.

In order to compare a variety of activation functions with different effective arities,
we need to take fair measures to ensure that the number of inputs needed to compute each output can be composed in a hidden layer sequence.
For an activation function with effective arity $\nAry$, the highest number of input arguments that could have been considered by hidden layer $\ell$ is $\nAry^\ell$.
Thus, hidden layer $\ell$ must have at least $32 \lceil \frac{\gamma}{\nAry^\ell} \rceil$ elements and, 
counting the final layer, we need $L = \lceil \log_\nAry (\gamma) \rceil$ layers to obtain outputs with enough computational pathways to at least match the number of input dependencies that are actually present in the ground truth.
 
Despite taking these measures, we know that such compositions are not capable of representing the complete variety of higher-arity truth functions,
which is one advantage of simply using a higher-arity layer.
For example, there are $2^{2^4} = 65,536$ distinct quaternary truth functions.
If we attempt to replicate these using only compositions of 3 binary operations (denoted $\circ_1$, $\circ_2$, and $\circ_3$) of the form
\small
\begin{align*}
&(\vX_1 \circ_1 \vX_2) \circ_3 (\vX_3 \circ_2 \vX_4),\quad
(\vX_1 \circ_1 \vX_3) \circ_3 (\vX_2 \circ_2 \vX_4),\\
&\quad\text{or}\quad (\vX_1 \circ_1 \vX_4) \circ_3 (\vX_2 \circ_2 \vX_3),
\end{align*}
\normalsize
we will find there are only 1,208 distinct functions.
Thus, both this counting argument and our experimental results, shown in \Cref{fig:logic-mlp-npar},
demonstrate that simple compositions of low-arity activation functions do not always have sufficient flexibility to reconstruct high-arity logic.

\begin{figure}[h!]
	\centering
	\includegraphics[width=0.47\textwidth]{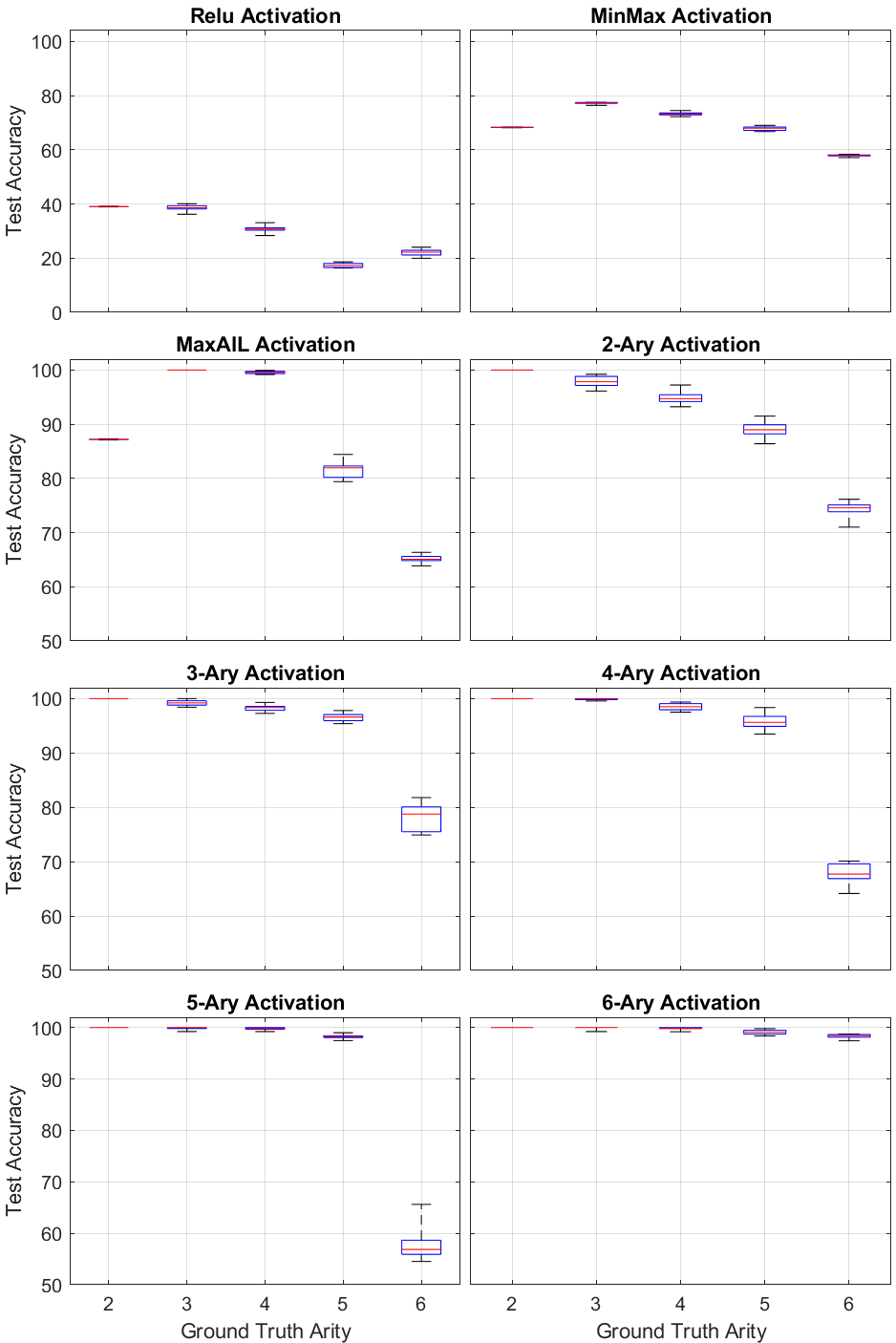}
	\caption{
	Activation function comparison using several ground-truth arities.
	ReLU performs significantly worse; note the wider accuracy scale used in the first row.
	MinMax activation performs better, but still worse than the other activation functions we compared.
	2-layer compositions of MaxAIL perform well up to the 4-ary ground truth,
	but our activation functions achieve nearly perfect accuracy, despite brief training, on matched and lower ground-truth arities.
	}
  \label{fig:logic-mlp}
\end{figure}

Each set of comparisons is performed for a range of ground-truth arities, $\nAry = 2, 3, \ldots, 6$.
Training, validation, and test datasets contain 5000, 2500, and 2500 cases, respectively.
Every ground truth arity uses the same training, validation, and test sets for all activation functions and all training trials.
To support reproducibility and illuminate the variety of outcomes,
we run 12 trials for each paired ground-truth and activation function.
Every trial begins by seeding the random number generator with the trial index before initializing network parameters and beginning training.
Training proceeds with 10 epochs of ADAM optimization \cite{Kingma2014} and
the epoch with optimal validation loss is saved.
\Cref{fig:logic-mlp} shows the range of test results from the cross-validation optimum.

Since most parameters in each network are contained within the linear transformation matrices,
the number of parameters scales roughly linearly with $\nAry$. See \Cref{sec:discussion}.

Because we know that the ground truth is generated by selecting only 1 out of 32 inputs per antecedent,
we would like regularization to compel most of the linear coefficients to zero.
To do this, we adaptively adjust $L_1$ regularization to cancel a small fraction of the near-largest gradient.
Specifically, we sort the average gradient magnitudes computed from ADAM and select the $15/16$-largest element from the array.
This allows us to set the $L_1$ weight to cancel a small fraction, $0.1$, times this magnitude,
thus leaving the largest gradients nearly unchanged while pushing parameters with much smaller gradients to zero.
Results are plotted in \Cref{fig:logic-mlp} for several activation functions.

\subsection{MNIST CNN}
\label{sub:mnist_cnn}

We developed this activation function with the intention of supporting more information-efficient processing pathways.
To that end, we notice that typical CNNs are structured so that adjacent strides overlap.
Using a filter of width $f$ with $c_{\text{in}}$ input channels and $c_{\text{out}}$ output channels,
each application of the filter has a computational complexity $c_{\text{out}} c_{\text{in}} f^2$.
If the layer has a spatial width of $x$ and we use a stride distance of $s$, then each filter must be applied $\nicefrac{x^2}{s^2}$ times, disregarding edge effects with appropriate padding.
If this is followed by a $2 \times 2$ max pooling operation, the number of spacial elements reduces by a factor of 4.
Thus, the total number of fused multiply-accumulations (FMAs) is $c_{\text{out}} c_{\text{in}} f^2 x^2 s^{-2}$.
Taking $f=3$ and $s=1$ results in $9 c_{\text{out}} c_{\text{in}} x^2$ operations per layer.

We would like to forge an approach where adjacent neurons are efficiently summarized, touching each only once with a small filter.
Our activation functions, however, also require $\nAry \times c_{\text{out}}$ antecedents to obtain $c_{\text{out}}$ consequent channels.
Thus, using $f=2$ and $s=2$ and dropping the max pooling operation, we obtain the same size of output, but with a computational factor of $\nAry c_{\text{out}} c_{\text{in}} x^2$.
Even though our activation functions also have $2^\nAry c_{\text{out}}$ parameters, the complexity is still dominated by the convolution, provided $\nAry \leq 6$.

\begin{figure}[h!]
	\centering
	\includegraphics[width=0.48\textwidth]{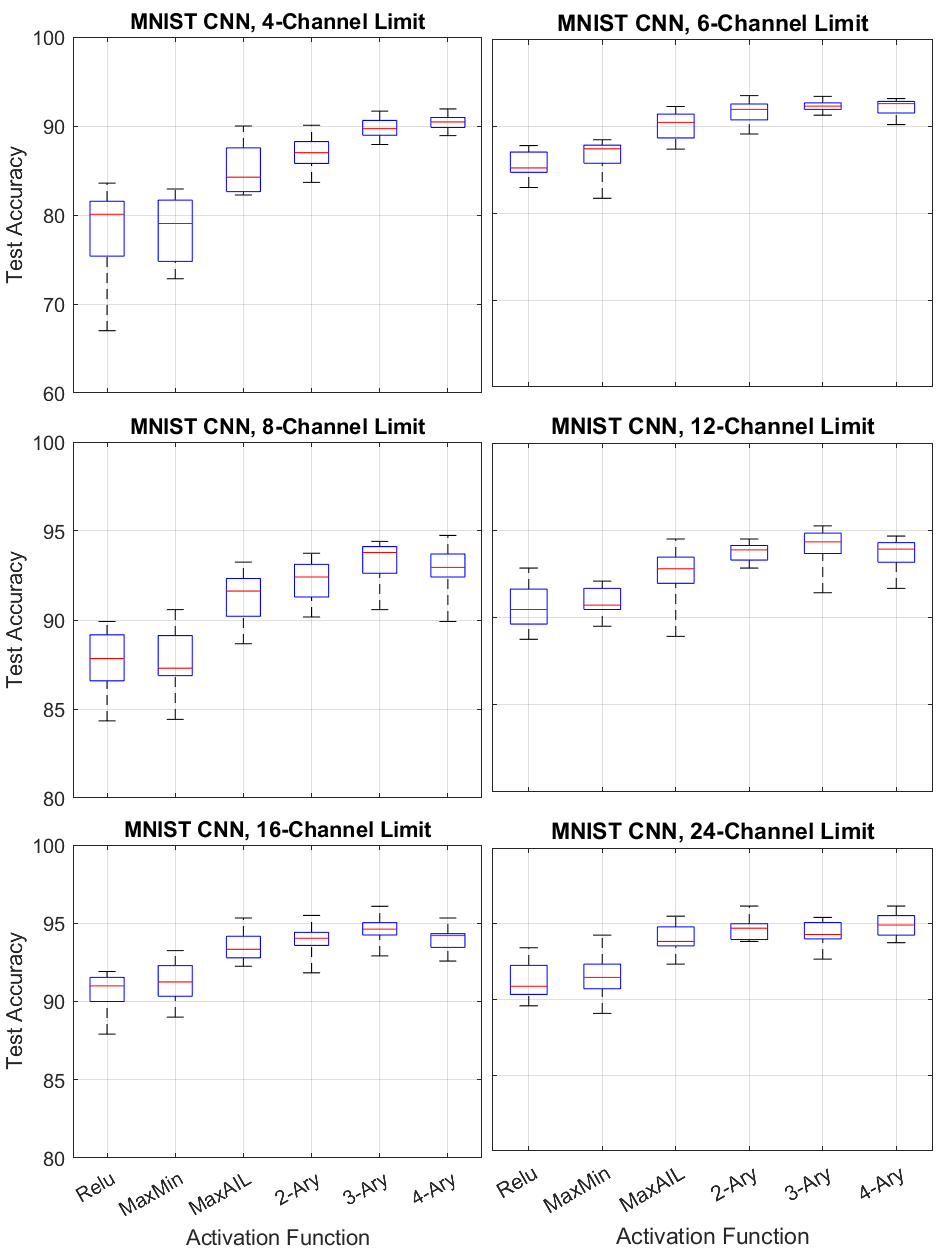}
	\caption{
	Activation function comparison using simple convolutional neural networks with several maximum channels.
	Both ReLU and MaxMin activation perform poorly for width-limited networks.
	MaxAIL is often competetive, but both ternary and quaternary functions perform well consistently and obtain superior accuracies when width is very limited.
	These results demonstrate the ability of higher-arity logic to use a limited number of channels more efficiently than competing approaches.
	}
  \label{fig:mnist-cnn}
\end{figure}

Since simple networks can obtain highly accurate predictions on the MNIST \cite{LeCun2010} dataset with only a fraction of examples,
each experiment only uses 6000 training images, 1200 validation images, and 1200 test images.
This is enough to resolve differences in expressivity for the activation functions we compare.
We determine the number of channels in each layer by multiplying the channels from the previous layer by 4 until we hit a maximum, $\wMax$.
Because we are interested in how efficiently higher-arity activation functions capture latent variables, we choose relatively low limits, $\wMax \in \{4, 6, 8, 12, 16, 24\}$. 
Note that 4 is the minimum number of Boolean variables that are capable of distinguishing 10 outcomes.
As before, we perform 12 trials, seeded by the trial index, using 10 epochs of ADAM.
The test set is evaluated from the cross-validation optimum and the distribution is plotted in \Cref{fig:mnist-cnn}.
Unsurprisingly, higher arity activation functions make the best use of width-restricted networks.

\subsection{CIFAR-10 CNN}
\label{sub:cifar10_cnn}

This set of experiments uses a standard CNN architecture for CIFAR-10 \cite{Krizhevsky2009} consisting of 3 convolutions, each with $3 \times 3$ filters, stride $1$, and interleaved by an ArgMaxAbs (AMA) layer,
\begin{align*}
\fAma(\mX) = \arg \max_{\vX_{ij}} \left| \vX_{ij} \right|.
\end{align*}
Note that for ReLU activation, AMA is identical to max-pooling, but it is more suitable for the other activation functions by preserving strong signed signals.
The first convolution computes $\wOne$ channels and the second and third compute $2 \wOne$ channels.
The last convolution is followed by a dense layer with $2 \wOne$ channels. 
We run tests for various widths, $\wOne \in \{16, 32, 64\}$. 
Finally, we have a dense layer to 10 softmax outputs.

\begin{figure}[h!]
	\centering
	\includegraphics[width=0.49\textwidth]{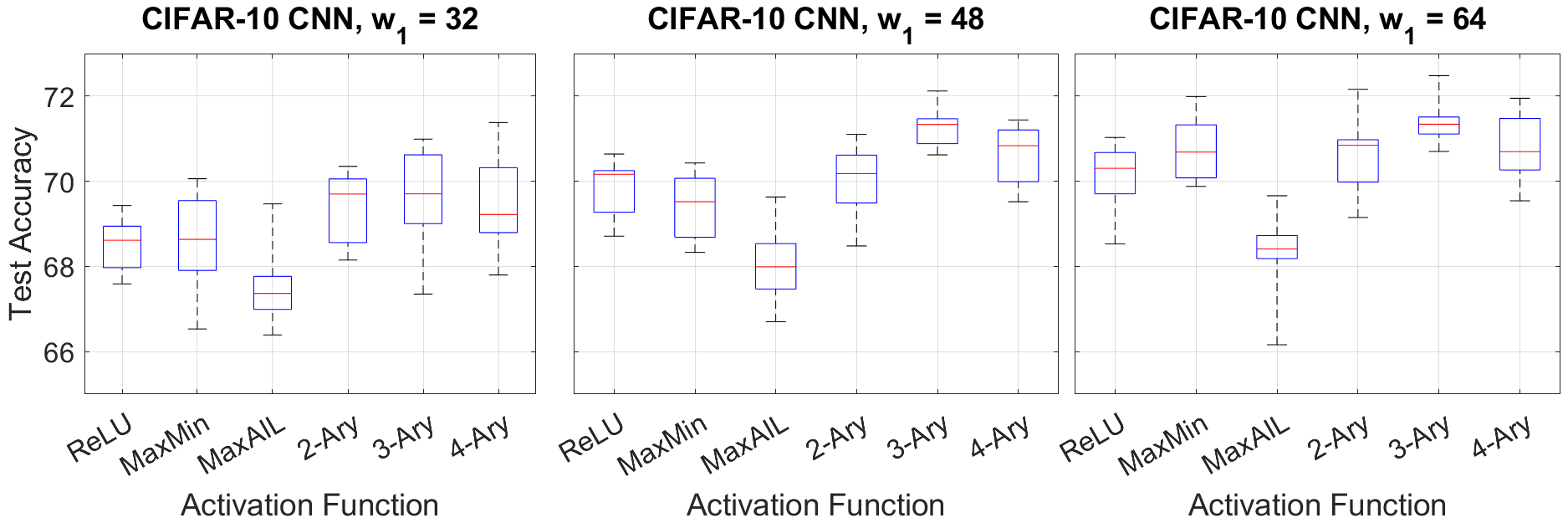}
	\caption{
	These test demonstrate that our adaptive logic activation functions can perform better than alternative activation functions in some cases.
	The best average results were obtained from ternary logic, yet we know that quaternary logic is capable of representing all the same outcomes.
	This suggests that efficient training methods that properly exploit sparsity remain to be understood.
	}
  \label{fig:cifar-cnn}
\end{figure}

Each experiment uses 40000 training images, 10000 validation images, and 10000 test images.
We use a fixed $L_1$ regularization weight, $w_1 = 0.5$, as well as a fixed $L_2$ weight, $w_2 = 0.01$, for all experiments.
Again, we perform 12 trials, seeded by the trial index, using 10 epochs of ADAM.
The test set is evaluated from the cross-validation optimum and the distribution is plotted in \Cref{fig:cifar-cnn}.
 
\section{Discussion and Summary}
\label{sec:discussion}

Since high-arity functions are capable of subsuming low-arity functions as well as ReLU and AIL activation functions,
we are convinced that the results we have obtained thus far do not demonstrate the full potential of high-arity logic.
The challenge remains to efficiently compel these functions to mimic simple logic when possible.
Doing so would allow us to use layers with a maximum expected arity and naturally reveal the latent complexity needed during training.

Solving this challenge will require more sophisticated machinery than an \textit{ad hoc} regularization strategy to deal with the increase in parameter dimensionality.
An $\nAry$-ary layer with $\wIn$ inputs and $\wOut$ outputs requires $\nAry \wOut \wIn$ parameters in the linear transformation, $\vY = \mM \vX$,
and $\wOut 2^\nAry$ activation parameters in $\mTheta$.
We expect the increase in parameters to be dominated by the matrix $\mM$ in practice.
For small $\nAry$ and even moderate widths, the number of parameters grows linearly with $\nAry$.
For example, taking $\wIn = \wOut = 32$ and $\nAry = 1, 2, 3, 4, 5,$ and $6$, the number of parameters in one layer grows as $P = 1088, 2176, 3328, 4608, 5120,$ and $6176$.
The exponential term only has a small impact until $\nAry > \log_2(\wIn)$.
Yet, even linear parameter growth can be problematic if we do not have a robust mechanism to limit parameter complexity as $\nAry$ increases.

Even without a method to efficiently handle increased parameters for higher-arity logic,
both binary and ternary activation functions still provide a useful increase in network expressivity.
Both of these activation functions consistently outperform Relu, MaxMin, and MaxAIL activation functions on the simple networks we tested.
Further, since our ternary activation functions open the ability to learn conditional reasoning in a single layer,
this simple change may open a new domain of learning capacity and improve our ability to extract complex predictions from data.

\subsection{Summary}

We introduced a framework to learn arbitrary belief functions that accept $n$ antecedent arguments.
This allows training to adjust the underlying logic to not only coherently account for antecedent uncertainty, but also each conditional consequent uncertainty,
which is more consistent with Bayesian reasoning.
By representing all probabilities (antecedents, belief table conditional probabilities, and consequents) as logits, we are able to systematically derive computationally efficient approximations.
Moreover, the resulting gradients are well-suited to optimization, because backpropagation transmits full gradient magnitudes to relevant antecedents and belief table parameters.
We also showed how a particular basis allows sparse representations to replicate lower arity logic using the same number of nonzeros as if we had used a lower arity activation function.

Our approach allows dimensionality to work in our favor.
Although implementing $\nAry$ argument logic requires $2^\nAry$ parameters per activation channel,
the resulting function becomes capable of distinguishing $3^{2^\nAry}$ qualitatively distinct functions,
thus dramatically increasing the ability of an architecture to discover useful predictions and avoiding relying on complicated compositions that may be difficult to adjust in order to correct isolated predictions.

We showed that our activation functions are capable of learning random truth functions of 6 arguments within a single layer.
Our experiments with MNIST and CIFAR-10 datasets show that even just ternary logic may provide a useful increase in expressivity over alternative activation functions.
Given that Boolean logic provides a mathematically rigorous framework for analyzing the relationships between propositions,
we believe that a soft logic framework, generalizing binary truth function to $\nAry$-ary belief functions, will prove to be a powerful tool in abstract learning algorithms.
Since our activation functions associate sparsity with logical complexity, this provides an important pathway to suppress logical complexity during training,
which we believe has a fundamental relationship with rigorously justified prior belief and few-shot learning.

\section*{Acknowledgements}

We extend our sincere appreciation to David Kavaler for providing feedback on an early draft of this manuscript.

\begin{flushleft}
\footnotesize\strut
This work was funded by the U.S.~Department of Energy and the Laboratory Directed Research and Development program at Sandia National Laboratories.\\ \vspace{2mm}

Sandia National Laboratories is a multimission laboratory managed and operated by National Technology and Engineering Solutions of Sandia, LLC.,
a wholly owned subsidiary of Honeywell International, Inc., for the U.S. Department of Energy's National Nuclear Security Administration under contract
DE-NA-0003525. This paper describes objective technical results and analysis. Any subjective views or opinions that might be expressed in the paper
do not necessarily represent the views of the U.S. Department of Energy or the United States Government.
\end{flushleft}

\appendix
\section*{Proof of Irrelevant Zeros}
\label{sec:appendix}
We begin by considering the last antecedent.
We can partition the truth table as $\vA_k = [\vAlpha_0 \quad \vAlpha_1]$, so that $\vAlpha_0 \in \reals^{2^{\nAry-1}}$ and $\vAlpha_1 \in \reals^{2^{\nAry-1}}$ give consequent logits when the last antecedent is false and true, respectively.
Since $\vA_k = \vTheta_k \mB$, we can also partition the parameters using matching sizes as $\vTheta_k = [\vPhi_0 \quad \vPhi_1]$.
Examining $\mB$ easily shows $\vAlpha_0 = \vPhi_0 - \vPhi_1$ and $\vAlpha_1 = \vPhi_0 + \vPhi_1$.
From the definition of irrelevant antecedents, the last antecedent is irrelevant if and only if $\vAlpha_0 = \vAlpha_1$.
It immediately follows that this is equivalent to $\vPhi_1 = 0$, i.e.~all columns indexed by $j$ for which $\text{bit}_\nAry(j)=1$.

We can apply this result to any antecedent by permuting the order of antecedents as follows.
Let $\text{permute}(\cdot)$ be the desired permutation on $\nAry$-bit sequences. 
We can form a permutation matrix $\mP$ that maps all $2^\nAry$ original columns, indexed by $j$, to permuted columns, indexed by $j'$, as $\mP_{jj'}=1$ for all $\text{bits}(j') = \text{permute}(\text{bits}(j))$.
Denoting all permuted representations using primes, we have
\begin{align*}
\vA'_k = \vA_k \mP = \left[ \vTheta_k \mP \right] \left[ \mP^T \mB \mP \right] = \vTheta_k' \mB'. 
\end{align*}
Since the structure of the Kronecker product for $\mB$ gives the element wise formula
$\mB_{\ell j} = (-1)^{\lor_{i=1}^\nAry ( \text{bit}_i(\ell) \land \lnot\text{bit}_i(j) )}$,
applying the same permutation to rows and columns of $\mB$ merely changes the order of $\fOr$ compositions in the exponent, so $\mB' = \mB$.
Thus, any antecedent reordering applied as a permutation to $\vA_k$ is represented by the same permutation to $\vTheta_k$.

If $\bY_{ki} \in \mathcal{I}$, permuting $i \leftrightarrow \nAry$ implies that $\vTheta'_{kj'}=0$ for all permuted columns, index by $j'$, with $\text{bit}_\nAry(j')=1$,
which means in the original order $\vTheta_{kj} = 0$ for all $\text{bit}_i(j)=1$.

\section*{Network Parameters}
\begin{table}[h]
\centering
\caption{Number of Parameters in Logic Networks}
\begin{tabular}{| r | r | r | r | r | r | r |} \cline{1-7}
\multicolumn{2}{| r |}{Arity} &    2 &     3 &     4 &     5 &     6  \\ \hline 
\multirow{8}{*}{\rotatebox[origin=c]{90}{Activation Function}}
&    Relu &  1024 &  4096 &  4096 & 11264 & 11264  \\ \cline{2-7}
&  MinMax &  1024 &  4096 &  4096 & 11264 & 11264  \\ \cline{2-7}
&  MaxAIL &  2048 &  8192 &  8192 & 22528 & 22528  \\ \cline{2-7}
&   2-Ary &  2176 &  8576 &  8576 & 23296 & 23296  \\ \cline{2-7}
&   3-Ary &  3328 &  3328 & 13056 & 13056 & 13056  \\ \cline{2-7}
&   4-Ary &  4608 &  4608 &  4608 & 17920 & 17920  \\ \cline{2-7}
&   5-Ary &  6144 &  6144 &  6144 &  6144 & 23552  \\ \cline{2-7}
&   6-Ary &  8192 &  8192 &  8192 &  8192 &  8192  \\ \hline 
\end{tabular}
\label{fig:logic-mlp-npar}
\end{table}

\begin{table}[h]
\centering
\caption{Number of Parameters in MNIST CNNs}
\begin{tabular}{| r | r | r | r | r | r | r | r |} \cline{1-8}
\multicolumn{2}{| r |}{Channel Limit} &     4 &     6 &     8 &    12 &    16 &    24  \\ \hline 
\multirow{6}{*}{\rotatebox[origin=c]{90}{Activation}}
&    Relu &   312 &   604 &   992 &  2056 &  3504 &  6656  \\ \cline{2-8}
&  MaxMin &   312 &   604 &   992 &  2056 &  3504 &  6656  \\ \cline{2-8}
&  MaxAIL &   584 &  1148 &  1904 &  3992 &  6848 & 13072  \\ \cline{2-8}
&   2-Ary &   664 &  1260 &  2048 &  4200 &  7120 & 13440  \\ \cline{2-8}
&   3-Ary &  1016 &  1916 &  3104 &  6344 & 10736 & 20224  \\ \cline{2-8}
&   4-Ary &  1448 &  2684 &  4304 &  8696 & 14624 & 27376  \\ \hline 
\end{tabular}
\label{fig:mnist-cnn-npar}
\end{table}

\begin{table}[h]
\centering
\caption{Number of Parameters in CIFAR-10 CNNs}
\begin{tabular}{| r | r | r | r | r |} \cline{1-5}
\multicolumn{2}{| r |}{First Conv.~Width} &    16 &    24 &    32  \\ \hline 
\multirow{6}{*}{\rotatebox[origin=c]{90}{Activation}}
&    ReLU & 80112 & 179688 & 318944  \\ \cline{2-5}
&  MaxMin & 80112 & 179688 & 318944  \\ \cline{2-5}
&  MaxAIL & 159904 & 358896 & 637248  \\ \cline{2-5}
&   2-Ary & 160352 & 359568 & 638144  \\ \cline{2-5}
&   3-Ary & 240592 & 539448 & 957344  \\ \cline{2-5}
&   4-Ary & 321280 & 720000 & 1277440  \\ \hline 
\end{tabular}
\label{fig:cifar10-cnn-npar}
\end{table}

\begin{IEEEbiography}[{\includegraphics[width=1in,height=1.25in,clip,keepaspectratio]{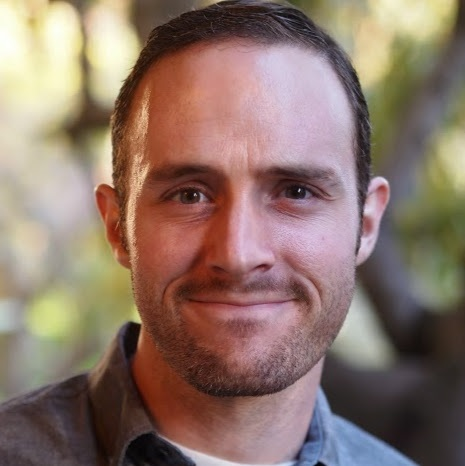}}]{Jed Duersch}
is a Principal Investigator at Sandia National Laboratories studying the relationship between complexity suppression and justified uncertainty quantification in machine learning algorithms.
He received B.A.~degrees in Mathematics and Physics as well as an M.A.~in Physics and Ph.D.~in Applied Mathematics from the University of California, Berkeley.
His research interests include information theory, Bayesian inference, algorithmic probability, dimensionality reduction, tensor decomposition, and efficient numerical methods.
\end{IEEEbiography}

\begin{IEEEbiography}[{\includegraphics[width=1in,height=1.25in,clip,keepaspectratio]{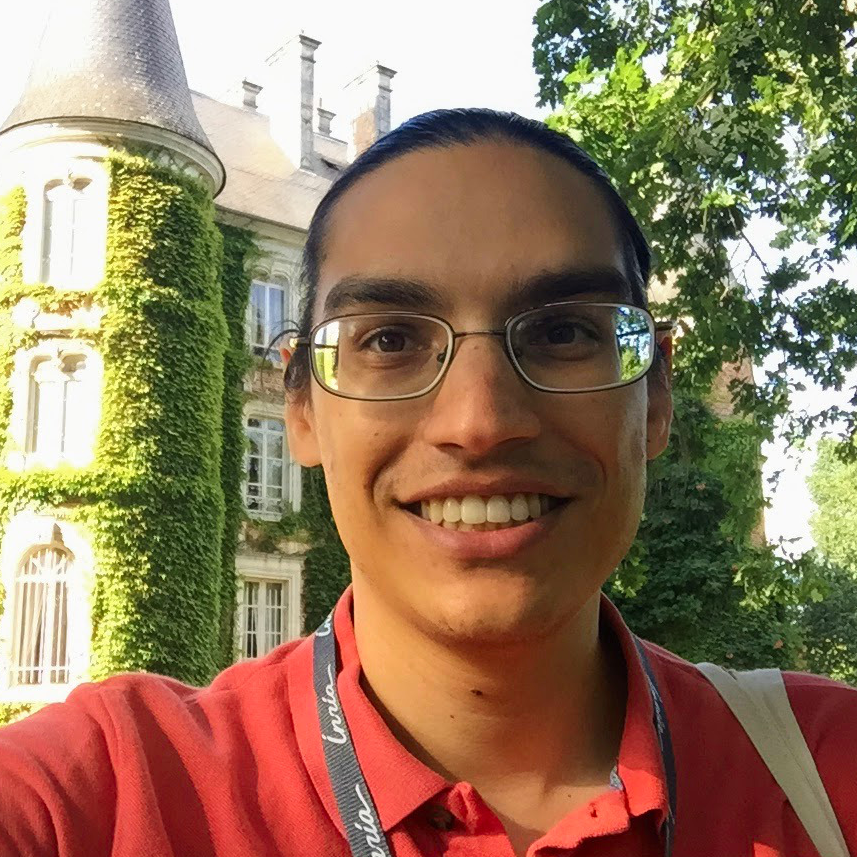}}]{Thomas Catanach}
is a senior member of the technical staff at Sandia National Laboratories.
Prior to Sandia, he received his Ph.D.~in Applied and Computational Mathematics from the California Institute of Technology and a B.S.~in Physics from the University of Notre Dame.
His research focuses on the theory, algorithms, and application of Bayesian inference, particularly for uncertainty quantification and machine learning.
He has worked in a variety of application areas including geophysics, structural engineering, power systems, and synthetic biology.
\end{IEEEbiography}

\begin{IEEEbiography}[{\includegraphics[width=1in,height=1.25in,clip,keepaspectratio]{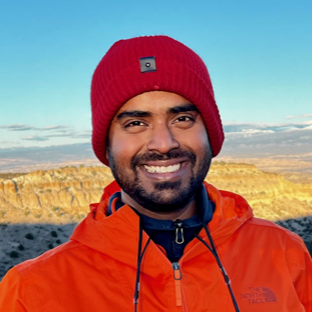}}]{Niladri Das}
received the B.Eng.~degree from Jadavpur University, West Bengal, India, in 2012,
the M.Tech.~degree in 2014 from Indian Institute of Technology Kanpur,
India, and his Ph.D.~degree from Texas A\&M University, Texas, USA in 2020.
He is currently a postdoctoral appointee at Sandia National Laboratories, California, USA, where he is currently working on machine learning and Bayesian inference.
\end{IEEEbiography}

\vfill

\bibliographystyle{IEEEtran}
\bibliography{IEEEabrv,soft_logic_arxiv_v1}

\end{document}